\documentclass{raw/tADR2e}

\usepackage[hyphens]{url} 
\usepackage{balance} 
\usepackage{comment} 
\usepackage{mathtools} 

\DeclareMathOperator*{\argmax}{argmax}

\usepackage{color}    

\begin{document}

\jvol{00}
\jnum{00}
\jyear{2020}
\jmonth{Xxxx}

\articletype{DRAFT FOR LIMITED CIRCULATION}



\title{
    Map completion from partial observation\\using the global structure of multiple environmental maps
}
    
\author{
    Yuki Katsumata$^{a}$,
    Akinori Kanechika$^{a}$,
    Akira Taniguchi$^{a}$$^{\ast}$\thanks{$^\ast$Corresponding author. Email: a.taniguchi@em.ci.ritsumei.ac.jp\vspace{6pt}},
    Lotfi El Hafi$^{b}$,
    Yoshinobu Hagiwara$^{a}$,
    and Tadahiro Taniguchi$^{a}$\\
    \vspace{6pt}
    $^{a}${\em{College of Information Science and Engineering, Ritsumeikan University;\\
    1-1-1 Noji-Higashi, Kusatsu, Shiga, Japan}}\\
    $^{b}${\em{Ritsumeikan Global Innovation Research Organization, Ritsumeikan University;\\
    1-1-1 Noji-Higashi, Kusatsu, Shiga, Japan}}\\
    \vspace{6pt}
    \received{v2.0 released August 2021}
}

\maketitle



\begin{abstract}
    Using the spatial structure of various indoor environments as prior knowledge, the robot would construct the map more efficiently.
    Autonomous mobile robots generally apply simultaneous localization and mapping (SLAM) methods to understand the reachable area in newly visited environments.
    However, conventional mapping approaches are limited by only considering sensor observation and control signals to estimate the current environment map.
    This paper proposes a novel SLAM method, map completion network-based SLAM (MCN-SLAM), based on a probabilistic generative model incorporating deep neural networks for map completion.
    These map completion networks are primarily trained in the framework of generative adversarial networks (GANs) to extract the global structure of large amounts of existing map data.
    We show in experiments that the proposed method can estimate the environment map 1.3 times better than the previous SLAM methods in the situation of partial observation.
    \medskip
\end{abstract}



\section{Introduction}

Even for unobserved areas that could be predicted with universal common knowledge, such as a corridor or the corner of a room, it is necessary to get observations first to estimate the environment map in previous simultaneous localization and mapping (SLAM) research.
To understand the shape and structure of the reachable area in a newly visited environment, autonomous mobile robots generally use a SLAM method which consists of estimating the environment map and self-position concurrently~\cite{robotics, Whyte2006, Barley2006}.
For example, when an autonomous cleaning robot is introduced into a newly human living environment where the environmental map is unknown, the robot necessarily starts operation by acquiring observations using its own sensors to estimate the environment's reachable area.
However, in previous SLAM research, such as FastSLAM~\cite{Montemerlo2002, Montemerlo2003}, each cell of the occupancy grid map is estimated independently, and the estimation of the environment map is limited to observable areas only.
Therefore, to execute tasks using the entire environment map estimated in previous studies~\cite{Montemerlo2003, Kohlbrecher2011}, the robot needs to get sufficient observation beforehand.

In contrast, a human can predict the internal spatial structure of unknown areas to some extent, even if it does not reach everything inside the building, such as the corridor or the room next door.
We believe that this ability of humans comes from the prior learning of structures inside various buildings from their own experience.
In this study, the robot uses map completion in unknown environments to estimate the unobserved area from the partially observed area using the joint probability of all the occupancy grid map cells.
Map completion, which consists of estimating the entire environment map with sufficient environmental information from a partial environment map estimated using insufficient environmental information, can be thought of as an image completion task of computer vision.
In this regard, using generative adversarial networks (GANs)~\cite{Goodfellow2014} is now considered state of the art to perform image completion~\cite{Isola2017, Zheng2019}.
Indeed, GANs~\cite{Goodfellow2014, Salimans2016} have been attracting attention because they can generate genuine data compared to more conventional image generation methods~\cite{Karras2018, Karras2019}.
Therefore, we adopt the framework of GANs for map completion and integrate it with a traditional SLAM method.

\begin{figure}
    \begin{center}
        \includegraphics[width=0.65\linewidth]{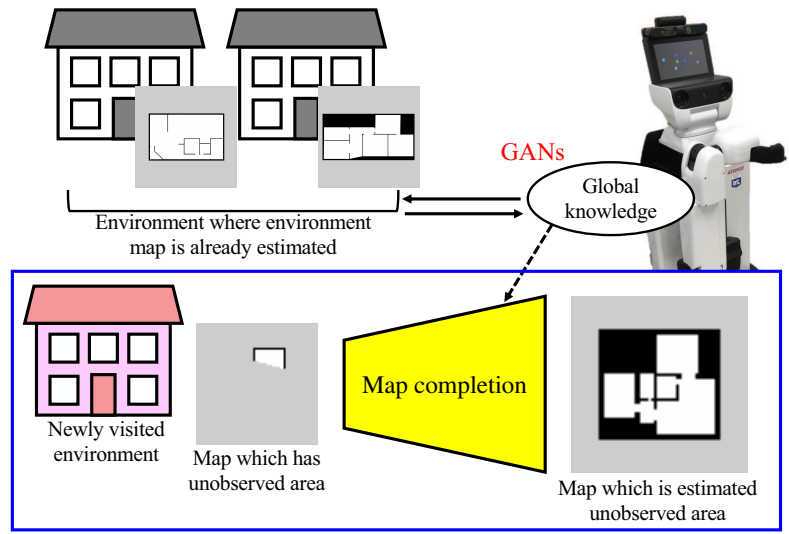}
        \caption{
            Overview of the proposed method: map completion network-based SLAM (MCN-SLAM).
            MCN-SLAM acquires knowledge of the global room structure from the known environments and transfers it to the unknown environment.
            Global knowledge includes features represented by occupancy grid maps, such as wall layout and room size. 
            The map is complemented by the interaction of global knowledge with partially generated maps using SLAM in the newly visited environment.
        }
        \label{fig:slam:overview}
    \end{center}
\end{figure}

We propose a novel SLAM method, map completion network-based SLAM (MCN-SLAM), that uses map completion trained in the framework of a generative model, and incorporates the prior distribution of the map that modeled the joint distribution of the occupancy probabilities of all the grid cells.
Figure~\ref{fig:slam:overview} shows an overview of the proposed MCN-SLAM.
With this method, we extract the global structure of maps of multiple environments and use it as a prior distribution for map completion.
Compared to previous SLAM methods where the map is generated only in one environment, MCN-SLAM can acquire or transfer the knowledge of the general room structure from multiple environments.
Hence, our contributions with MCN-SLAM are as follows:
\begin{enumerate}
    \item We extend the FastSLAM formulation by incorporating map completion using generative models.
    \item We show that our method can estimate the map of the unobserved area from the observed area with higher accuracy than the traditional SLAM methods.
    \item We show that our method can acquire and transfer general room structure knowledge from multiple environments.
\end{enumerate}

The remainder of the paper is structured as follows.
Section~\ref{realted_work} introduces the related work.
Section~\ref{proposed_method} describes the proposed method MCN-SLAM.
Section~\ref{experiment} compares the proposed method with previous related methods.
Finally, Section~\ref{conclusion} concludes this study with perspectives for future work.



\section{Related work} \label{realted_work}

This section introduces works related to the proposed method, which intersects two fields:
simultaneous localization and mapping (SLAM), and image completion with deep generative models (DGMs).


\subsection{Simultaneous localization and mapping (SLAM)}

SLAM consists of estimating the environment map and self-position at the same time~\cite{robotics, Whyte2006, Barley2006}.
Montemerer~et~al. proposed FastSLAM as an online method to solve SLAM~\cite{Montemerlo2002, Montemerlo2003} by using a Rao-Blackwellized particle filter.
Gmapping~\cite{ros_gmapping}, which is implemented as a standard Robot Operating System (ROS)~\cite{ros} package, is a grid-based version of FastSLAM.
The environment map estimated by Gmapping is in the form of an occupancy grid map and uses the mechanisms for handling occupancy grid maps prepared in ROS.

The occupancy grid map is one of the map representations that robots commonly use to accomplish various service tasks~\cite{robotics,lotfi2018WRS,ataniguchi2020spconavi,Katsumata2020SpCoMapGAN}.
It is a method that divides the environment into grid cells at fixed intervals and stores the occupancy probability of each cell in a 2D list format.
The cells with a high value of occupancy probability are assumed to be occupied, and the cells with a low value are assumed to be unoccupied and available for the robot to move to.
The occupancy grid map is more suitable for robot navigation tasks than other map representation methods because it can search for routes not occupied by obstacles on the map.
Assuming that the grid cell of the occupancy grid map with index $i$ is $g_i$, the occupancy grid map $m$ is the space divided by grid cells such that $m = \{ g_i \} \, (i \in S)$ where $S$ is the number of cells in the occupancy grid map.
A binary occupancy value is assigned to each grid $g_i$: $g_i = 1$ when occupied and $g_i = 0$ when unoccupied.
An occupancy grid map is used to expand an uncertain or unspecified region as an unsearched area.
The unsearched area is a cell for which it is impossible to determine its occupancy from the sensor values.

CNN-SLAM is a method that utilizes convolutional neural networks (CNN) to improve SLAM accuracy with depth prediction~\cite{Tateno2017}.
While CNN-SLAM is a method that uses neural networks to complement the observation, our proposed method uses neural networks to complement the environment map estimated from the observation.

There are also methods to complement the unobserved parts of the environment map using the information of the layout of the rooms~\cite{Liu2014, Luperto2019_springer, Luperto2019}.
In particular, Luperto~et~al. proposed a method that identifies the layout of a partially known room from the walls on the 2D grid map and estimates the room layout from the known parts of the environment to the unknown parts by propagating the regularities~\cite{Luperto2019}.
However, map completion functions as post-processing of SLAM and is limited to partial area completion.
In our study, SLAM and map completion are theoretically integrated as one probabilistic generative model, and maps are sequentially generated from the global knowledge of spatial structures.


\subsection{Image completion with deep generative models (DGMs)}

When performing map completion, we want the robot to maintain some features from the observation to the semantic map, such as the occupancy grid and the unoccupied area estimated from the observations in the occupancy grid map.
Therefore, map completion can be solved in the pix2pix~\cite{Isola2017} framework, which is one approach of GANs~\cite{Goodfellow2014}, by regarding the occupancy grid map completion as a similar task of image completion.
Indeed pix2pix, which learns the relationship between a pair of images using U-net~\cite{Ronneberger2015} in the generator part, enables image completion and line drawing coloring.
pix2pix is a method based on the conditional GAN (CGAN) framework, which is a GAN method using conditional information to learn the relationships between the training data and the condition data~\cite{Mirza2014}.
The pix2pix's generator has skip connections to extract the features of the original image at the encoder by using the image as input and adding these features at the decoder to the output of the generator.

GANs~\cite{Goodfellow2014} are generative models that simultaneously learn two networks: the generator that performs data generation and the discriminator that estimates the distance between the distribution of the data generated by the generator and that of the real data.
GANs have attracted increasing attention because they can generate genuine data compared to conventional generative models~\cite{Karras2018, Karras2019}.
However, GANs face the problem of difficult convergence because two networks are trained simultaneously~\cite{Salimans2016}.
In this regard, Miyata~et~al. proposed spectral normalization GAN (SNGAN) with improved performance using spectral normalization as the discriminator weight~\cite{Miyato2018}.
Our study uses SNGAN to help the learning converge because SNGAN is easy to implement and can be combined with other GAN structures.
For the above reasons, we adopt U-net for the generator and pix2pix and SNGAN for the discriminator.

Katyal~et~al. proposed an active SLAM method that combines map estimation and map completion using U-net to perform map completion and use an ambiguous position where the map cannot be complemented as the next search position~\cite{Katyal2019}.
Shrestha~et~al. proposed a variational autoencoder (VAE) deep neural network that learns to predict unseen regions of building floor plans and used it for path planning to enhance exploration performance~\cite{Shrestha2019}.
These methods use local maps to improve the estimation accuracy and determine the navigation targets.
In contrast, our study attempts to estimate the shape of the global environment.



\section{Proposed Method: MCN-SLAM} \label{proposed_method}

\begin{figure}
    \begin{center}
        \includegraphics[width=0.8\linewidth]{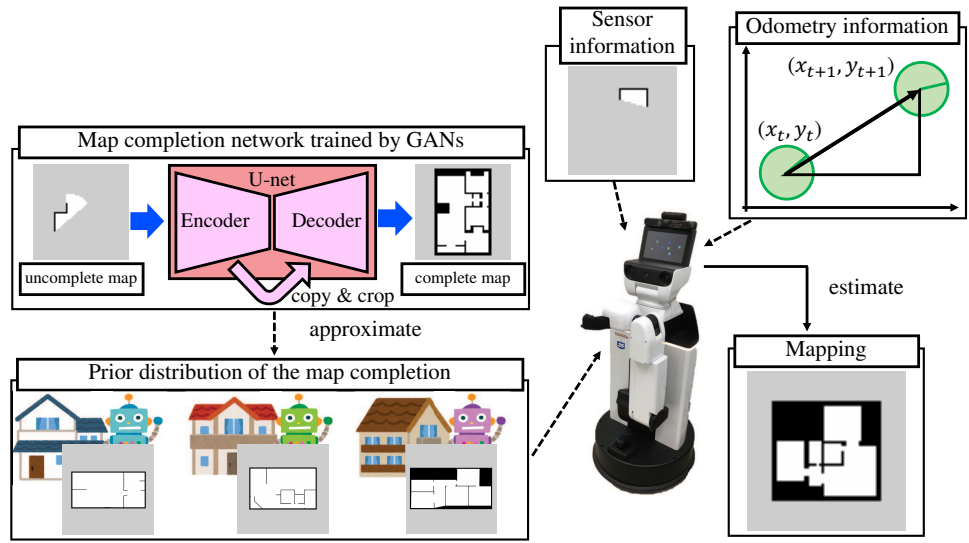}
        \caption{
            The proposed method, map completion network-based SLAM (MCN-SLAM), trains a network to model the global structure extracted from multiple environmental maps using GANs and estimates the unobserved areas based on partial observation.
            MCN-SLAM extracts the global structure of maps of multiple environments as a prior distribution for map completion.
            In a newly visited environment, the robot estimates the map using that distribution and observation information.
            Sensor information and odometry information are used to construct the uncomplete map.
            The complete map is obtained as a sample from the conditional distribution. 
        }
        \label{fig:slam:gaiyou}
    \end{center}
\end{figure}

We propose a map completion network-based SLAM (MCN-SLAM) that estimates the unobserved areas based on partial observation by using the global structure extracted from multiple environmental maps.
In this method, a network for map completion is trained in advance using the architecture of a generative model, and the prior distribution of the map is modeled by the joint distribution of the occupancy probabilities of all cells.
An overview of the proposed MCN-SLAM method is shown in Figure~\ref{fig:slam:gaiyou}.

\begin{figure}
    \begin{center}
        \includegraphics[width=0.5\linewidth]{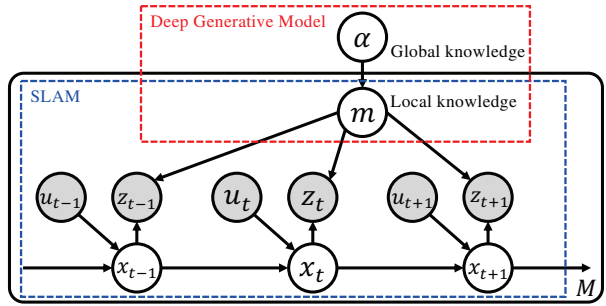}
        \caption{
            Graphical model representation of the MCN-SLAM generative process where the gray nodes indicate observation variables and the white nodes unobserved variables.
            The part surrounded by the blue dotted frame is the model representation of the SLAM of one environment.
            The global knowledge is assumed as a parameter of the prior distribution for maps that integrate local knowledge in various environments.
        }
        \label{fig:slam:graphicalmodel}
    \end{center}
\end{figure}

In addition, Figure~\ref{fig:slam:graphicalmodel} shows the graphical model representation of MCN-SLAM, and Table~\ref{tbl:slam:graphicalmodel} its variables.
Equations (\ref{equ:slam:gen_x}) -- (\ref{equ:slam:gen_m}) describe the generative process of the proposed graphical model as follows:
\begin{align}
    x_t &\sim p(x_t \mid x_{t-1}, u_t),\label{equ:slam:gen_x}\\
    z_t &\sim p(z_t \mid x_t, m),\label{equ:slam:gen_z}\\
    m &\sim p(m \mid \alpha),\label{equ:slam:gen_m}
\end{align}
where the probability distribution of Equation~(\ref{equ:slam:gen_x}) represents a motion model in SLAM, i.e. a state transition model, and the probability distribution of Equation~(\ref{equ:slam:gen_z}) a measurement model in SLAM.

In the following, Section~\ref{sec:slam:training} describes the training of the map completion network, and Section~\ref{sec:slam:slam} the algorithm that performs SLAM.

\begin{table}
    \tbl{
        Definition of the variables in the graphical model of MCN-SLAM.
    }{
        \begin{tabular}{cc}
            \toprule
            \textbf{Symbol} & \textbf{Definition}\\
            \colrule
            $m$ & Map of the environment\\
            $x_t$ & Self-position of the robot\\
            $u_t$ & Control observation\\
            $z_t$ & Distance observation\\
            $\alpha$ & Parameter of the prior distribution of the map\\
            $M$ & Number of environments\\
            \botrule
        \end{tabular}
    }
    \label{tbl:slam:graphicalmodel}
\end{table}


\subsection{SLAM with map completion}
\label{sec:slam:slam}

The proposed MCN-SLAM method is an extension of FastSLAM.
FastSLAM is formalized as calculating the joint distribution of the trajectory of the self-positions $x_{0:t}$ and the map $m$.
Likewise, the formulation of MCN-SLAM is as follows:
\begin{align}
    p(x_{0:t}, m \mid z_{1:t}, u_{1:t}, \alpha) &= p(m \mid x_{0:t}, z_{1:t}, \alpha) p(x_{0:t} \mid z_{1:t}, u_{1:t}, \alpha)\nonumber\\
    &\approx \underbrace{p(m \mid x_{0:t}, z_{1:t}, \alpha)}_{\text{mapping with completion}} \underbrace{p(x_{0:t} \mid z_{1:t}, u_{1:t})}_{\substack{\text{self-localization} \\ \text{by particle filter}}},
    \label{equ:slam:slam1}
\end{align}
{where it is assumed that the parameter of the prior distribution {$\alpha$} is learned in advance.
The learning method is described in Section~\ref{sec:slam:training}.}

From the standpoint of generative models, the mapping process in SLAM is the inference of the posterior distribution of a map $m$, as shown in the first term of the Equation~(\ref{equ:slam:slam1}).
In the second term of Equation~(\ref{equ:slam:slam1}), the prior parameter $\alpha$ can be ignored as an approximation because the information outside the sensor observation has little effect on the self-localization, and self-localization is determined by a particle filter.
The first term of Equation~(\ref{equ:slam:slam1}) is the posterior probability distribution calculated from the information up to the present and the parameters that determine the shape of the prior distribution of the map.

Here it is difficult to estimate the shape of the distribution that models the joint distribution of the occupancy probabilities of all cells used as the prior distribution of the map.
Therefore, in this study, we approximate the distribution of data sampled from the prior distribution of the map by the distribution of data generated by the map completion network.


\subsection{Training map completion networks}
\label{sec:slam:training}

In this study, the dependency of the occupancy probability of the cells in the occupancy grid map is obtained from the estimation result of the map in the known environment. The map in an unknown environment is complemented by using the prior distribution of the occupancy grid map.

{The map $m_t$ estimated using the partial observation up to a certain time $t \leqq T$, where $T$ is the last time at which sufficient observation is given, is as follows:}
\begin{align}
    m_t &= \argmax_m p(m \mid x_{0:t}, z_{1:t}),\label{equ:slam:m_}
\end{align}
where $x_{0:t}$ uses the value with the maximum weight among the particles estimated by the particle filter.

The map completion task consists of estimating the environment map using $m_t$ and the parameter $\alpha$ that determines the shape of the prior distribution of the map.
{Since $m_t$ is a map of the environment estimated from $x_{0:t}$ and $z_{1:t}$, we approximate it by assuming that it has the information of $x_{0:t}$ and $z_{1:t}$.
This approximation is similar to the calculation process used for sequential updates of grid maps.
Therefore,} the first term in Equation~(\ref{equ:slam:slam1}) is approximated as follows:
\begin{align}
    p(m \mid x_{0:t}, z_{1:t}, \alpha) &\approx p(m \mid m_t, \alpha).
\end{align}
By approximating the probability distribution, various map complement candidates can be expressed stochastically.
Therefore, the completed environment map $m_s$ is sampled by the following:
\begin{align}
    m_s &\sim p(m \mid m_t, \alpha). \label{equ:slam:m_s}
\end{align}

To estimate the prior distribution for map completion, we consider the occupancy grid map as an image, train the network using an image completion method, and approximate the prior distribution by that network.
In this study, the prior distribution is approximated by U-net~\cite{Ronneberger2015}.
When training the map completion network, the distribution of the estimated environment map $m_s$ is trained to approximate the distribution of $m_T$.
Two types are prepared: with and without the discriminator loss in GANs.

\textbf{With the discriminator}:
The objective function $C$ for training the map completion network in the pix2pix framework, which is a CGAN~\cite{Mirza2014}, is as follows:
{\begin{align}
    C(G) &= - \log 4 + 2 {\mathcal{D}_{\rm JS}}[ p_{\rm data}(m \mid c) \| p_g(m \mid c) ],\label{equ:slam:CG}
\end{align}}
where $\mathcal{D}_{\rm JS}[ A \| B ]$ is the Jensen-Shannon divergence (JSD) between the distribution $A$ and $B$, $c$ the condition, $p_{\rm data}(\cdot)$ the training data distribution, and $p_g(\cdot)$ the distribution of the data generated by the generator $G$.
CGAN learns to estimate $p_g(m \mid c)$ that minimizes the distance from the distribution $p_{\rm data}(m \mid c)$ of the training data, which is the map $m_T$ estimated at the last time $T$, to minimize this objective function.

\textbf{Without the discriminator}:
It is also possible to train the generator $G$ directly without using the discriminator in GANs.
We use the L2 loss, i.e., the Euclidean distance, for training.
The objective function $C$ for training the map completion network using U-net is as follows:
\begin{align}
    C(G) &= {\mathcal{D}_{\rm L2}}[ m_T \| m_t ],\label{equ:slam:unet}
\end{align}
where ${\mathcal{D}_{\rm L2}}[ A \| B ]$ is the L2 loss function between the distributions $A$ and $B$.

In this study, since the condition is $m_t$, the right side of Equation~(\ref{equ:slam:m_s}) is expressed by the following equation using a network learned in the above framework:
\begin{align}
    p(m \mid m_t, \alpha) &\coloneqq p_g(m \mid c = m_t).\label{equ:slam:p_m}
\end{align}

Assuming that generating data from the network is sampling from the distribution of the generated data, the equation for sampling the completed environment map from the prior distribution and the map $m_t$ can be written from Equation~(\ref{equ:slam:p_m}) as follows:
\begin{align}
    m_s \sim p_g(m \mid c = m_t).\label{equ:slam:sampling}
\end{align}

\subsection{Processing of the input and output map images during training}
\textbf{Data format}: 
The occupancy grid map in Gmapping, which is used in the SLAM part of our proposed method, is a two-dimensional ${\rm height} \times {\rm width}$ tensor.
This map data is divided into the map image $m_{\rm image}$ and the mask image $m_{\rm mask}$, and is converted in a grayscale image.
The image data format of the occupancy grid map is $m_{\rm image} = \{ g_{{\rm image},i} \}$ $(i \in S)$, where $S$ represents the number of cells in the occupancy grid map.
Here, $g_{{\rm image},i}$ is the $i$-th pixel in which each cell of the occupancy grid map is described as unoccupied area, occupancy area, or unsearched area, with three values defined as follows:
\begin{align}
    g_{{\rm image},i} = \left\{
        \begin{array}{l}
            1.0 \quad (p(g_i) > 0.5){, \quad \text{occupancy area;}}\\
            0.5 \quad (p(g_i) = 0.5){, \quad \text{unsearched area;}}\\
            0.0 \quad (p(g_i) < 0.5){, \quad \text{unoccupied area.}}\\
        \end{array}
    \right.
    \label{eq:prob2image}
\end{align}
In addition, the mask image data in the unsearched area is {$m_{\rm mask} = \{ g_{{\rm mask},i} \}$ $(i \in S)$}.
Here, $g_{{\rm mask},i}$ is the image of each cell of the occupancy grid map {binarized} into a searched or an unsearched area as follows:
\begin{align}
    g_{{\rm mask},i} = \left\{
        \begin{array}{l}
            1.0 \quad (p(g_i) = 0.5)\\
            0.0 \quad (p(g_i) \neq 0.5)\\
        \end{array}.
    \right.
\end{align}

\textbf{Input and output of the generator}: 
The generator network receives two kind of images, $m_{\rm image}$ and $m_{\rm mask}$, and generates $m'_{\rm image} = \{ g'_{{\rm image},i} \}$ and $m'_{\rm mask} = \{ g'_{{\rm mask},i} \}$ correspondingly.
The input and output size is ${\rm height} \times {\rm width} \times 2$.

The outputs of the generator are converted into the image format of a ternary for the occupancy grid map.
The cell of the inputs $g_{{\rm image},i}$ and $g_{{\rm mask},i}$ are discrete values, but the cell of the outputs $g'_{{\rm image},i}$ and $g'_{{\rm mask},i}$ are continuous values because they pass through the network.
The process of converting the outputs of the generator, which are continuous values, into a discrete value is performed as follows:
\begin{align}
    g'_{{\rm converted\_image},i} &= \left\{
        \begin{array}{l}
            1.0 \quad (g'_{{\rm mask},i} < 0.5 \; {\rm and} \; g'_{{\rm image},i} \geq 0.5)\\
            0.5 \quad (g'_{{\rm mask},i} \geq 0.5)\\
            0.0 \quad (g'_{{\rm mask},i} < 0.5 \; {\rm and} \; g'_{{\rm image},i} < 0.5)\\
        \end{array}.
    \right.
\end{align}

In the continuous value representation of the occupancy grid map at the output of the generator, it is difficult to distinguish between the unsearched area and the searched area.
Therefore, $m'_{mask}$ is output to obtain information that supports the restoration of the map image by estimating the unsearched area.
Since U-net requires the input and output layers to have the same structure, mask images are also required for both the input and output regardless of the use of a discriminator.

\textbf{Input and output of the discriminator}: 
The inputs of the discriminator are $m_{\rm image}$, which is the map before completion, and $m'_{\rm converted\_image} = \{ g'_{{\rm converted\_image},i} \}$, which is the map after completion.
These inputs correspond to $m_{t}$ and $m_{T}$, respectively.
The output of the discriminator is a binary value, i.e., true or false.



\section{Experiment} \label{experiment}


\subsection{Experimental dataset}

We prepared two types of large-scale home environment datasets: the HouseExpo dataset~\cite{HouseExpo} and the HOME'S dataset~\cite{lifull}.
The HouseExpo dataset is a large-scale image dataset of 2D indoor layouts generated from the SUNCG dataset~\cite{suncg}.
In our experiments, we use the ground truth included in the dataset as the map data estimated using sufficient observation.
The map images generated by a Pseudo SLAM simulator are also included in the HouseExpo dataset, as the map data is estimated using insufficient observation.
We used 5,000 images as training data, 200 images as model validation data, and 100 images as test data.

On the other hand, the HOME'S dataset includes floor plans of apartments in Japan.
We use map images created by classifying the colored areas of the floor plans included in the dataset into the occupancy grid, unoccupied cells, and unsearched areas as the map data estimated using sufficient observation.
We also use map images created by replacing a part of these floor plans with an unsearched area as the map data estimated using insufficient observation.
We used 3,500 images as training data, 200 images as model validation data, and 100 images as test data.
{All the map images were resized to $64 \times 64$ pixels.}

\begin{figure}
    \begin{center}
        \includegraphics[width=8cm]{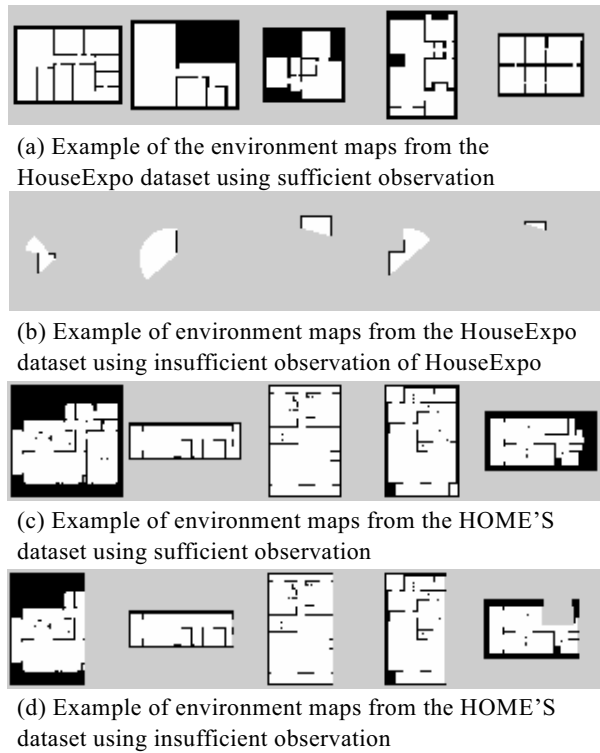}
        \caption{
            Examples of the dataset: (a) example of the environment maps from the HouseExpo dataset using sufficient observation, (b) example of environment maps from the HouseExpo dataset using insufficient observation of HouseExpo, (c) example of environment maps from the HOME'S dataset using sufficient observation, and (d) example of environment maps from the HOME'S dataset using insufficient observation.
        }
        \label{fig:slam:example_dataset}
    \end{center}
\end{figure}

Figure~\ref{fig:slam:example_dataset} shows some examples of the data.
As the HOME'S data are converted from the floor plans included in the dataset, they contain more noise compared to the HouseExpo dataset created in a simulator environment.
The environment maps of the HOME’S dataset using insufficient observation randomly lack 10--50\% of the map size of the dataset using sufficient observation.
The environment maps of the HouseExpo dataset using insufficient observation are generated by the first iteration of SLAM in the simulator.

\subsection{Network training}

We train a map completion network in the generative model framework.
The network architecture is based on U-net~\cite{Ronneberger2015} for the generator, similarly to pix2pix~\cite{Isola2017}.
For the map completion, we want to maintain some features from the encoder to the decoder, such as the structure of the part in the occupancy grid map estimated from observation and thus likely to be correct.
Therefore, we use skip connections in the generator to retain the information.

We also use spectral normalization~\cite{Miyato2018} at the discriminator of pix2pix~\cite{Isola2017} as it helps the convergence of GANs that are difficult to train.
In addition, spectral normalization can improve performance if added to the generator as well~\cite{Brock2018}.
In this study, we also adopt spectral normalization for U-net.

The network's training time was 14.5 h for 10,000 epochs when using an Intel Core i9 7980XE CPU combined with an Nvidia Quadro GV100 GPU.
The implementation was realized on Keras and TensorFlow.


\subsection{Experiment 1: Comparison of the inference models}
\label{sec:exp1}

We measure the accuracy of map completion using both the HouseExpo and HOME'S datasets.
The methods to be compared are as follows:
\begin{description}
    \item[(A)] MCN-SLAM (proposed): pix2pix (U-net+discriminator)
    \item[(B)] MCN-SLAM (proposed): U-net
    \item[(C)] Pseudo SLAM
    \item[(D)] Baseline
\end{description}
MCN-SLAM requires the skip connections of a U-net.
In this study, we train pix2pix, which is a combination of U-net and discriminator loss.
U-net, without discriminator loss, performs supervised learning using the training data.
Pseudo SLAM is an occupancy grid map image included in the HouseExpo dataset that was obtained from simulation.
Although the experimental results of Pseudo SLAM are not the results of the estimation by the actual robot, we consider the data obtained by the Pseudo SLAM simulator as the occupancy grid map estimated when a robot performs SLAM in the environment.
We use the training data with the highest precision with $m_s$ for the estimated map as the result of the baseline method.
This shows that the data output by the network does not simply mimic the training data but that the distribution of the training data is closer to the distribution of the generated data by the inference model.

In addition, to show that the proposed method can obtain the prior distribution corresponding to the domain of the training data, we evaluate the proposed method with the same index using another domain.
Here, we compare using the test data of the HouseExpo dataset for the network of map completion trained using the HouseExpo dataset and the HOME'S dataset.

In this experiment, we assume that the test data of the dataset to be $m_t$, and evaluate the accuracy of the map completion method by the proposed method and comparison methods.
We evaluate them in terms of accuracy, precision, recall, and F-measure.
In this case, we compare the correctness of an occupancy area that represents the shape of the room.
Therefore, we use a binary classification: is an occupancy area or is not.

\begin{table}
    \tbl{
        Experiment 1: Experimental results using the HouseExpo dataset as test data.
        The top two columns show the experimental results using HouseExpo for the training data, and the bottom two columns show the experimental results using HOME'S for the training data.
    }{
        \begin{tabular}{l|cccc}
            \toprule
            Method & Accuracy & Precision & Recall & F-measure\\
            \colrule
            (A) MCN-SLAM (proposed): pix2pix trained with HouseExpo & \underline{0.5831} & \underline{\bf 0.8613} & \underline{0.3762} & \underline{0.4939}\\
            (B) MCN-SLAM (proposed): U-net trained with HouseExpo & \underline{\bf 0.5852} & \underline{0.8612} & \underline{\bf 0.3807} & \underline{\bf 0.4963}\\
            (C) Pseudo SLAM & 0.4340 & 0.8517 & 0.0616 & 0.1110\\
            (D) Baseline & 0.4893 & 0.8026 & 0.1753 & 0.2734\\
            \colrule
            (A) MCN-SLAM (proposed): pix2pix trained with HOME'S & 0.5065 & 0.8537 & 0.2222 & 0.3481\\
            (B) MCN-SLAM (proposed): U-net trained with HOME'S & 0.5257 & 0.7915 & 0.2976 & 0.4286\\
            \botrule
        \end{tabular}
    }
    \label{tab:slam:exp1result}
\end{table}

\begin{figure}
    \begin{center}
        \includegraphics[width=10cm]{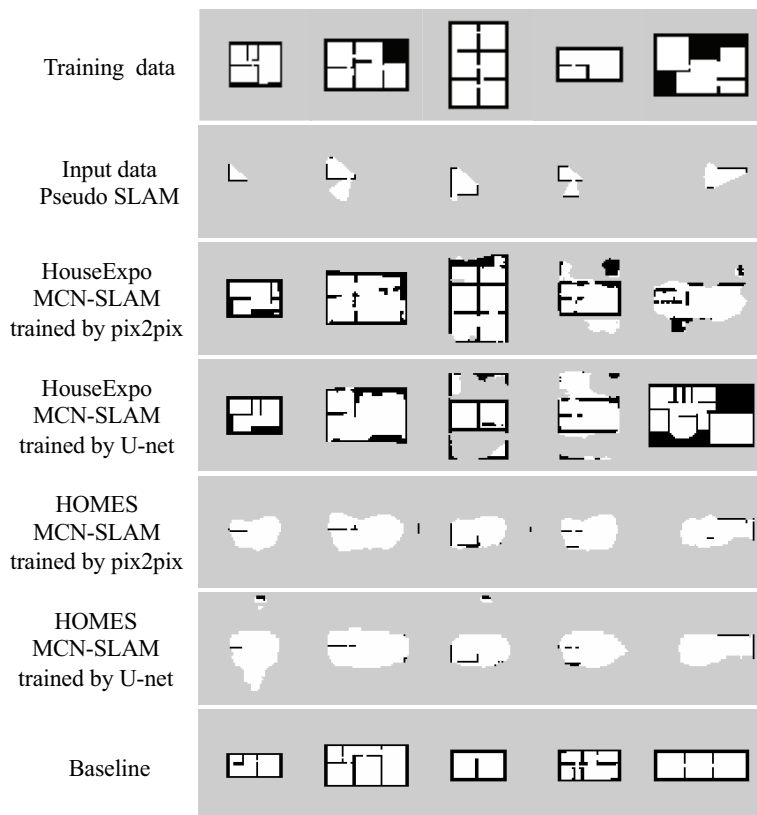}
        \caption{
            Experiment 1: Example of generated maps using the HouseExpo dataset as training data.
        }
        \label{fig:slam:exp1result}
    \end{center}
\end{figure}

Table~\ref{tab:slam:exp1result} shows that the proposed method performed better in accuracy and F-measure than the comparison methods, and Figure~\ref{fig:slam:exp1result} shows an example of generated map images.
This indicates that the network trained by the proposed method can perform map completion.
In addition, comparing pix2pix and U-net, the methods without discriminator performed better to complete maps.
{As a result, the completed map was made according to the shape of the partially observed map extended by the acquired knowledge.}
Furthermore, the comparison between the proposed method and the baseline method shows that the data output by the generator network does not simply mimic the training data but that the distribution of the training data is closer to the distribution of the data generated by the inference model.

When comparing the networks trained with the HouseExpo dataset and the networks trained with the HOME'S dataset, the accuracy and F-measure performed better on networks trained using the same HouseExpo dataset as the test data.
This indicates that the distribution that the map completion network can approximate is a domain-dependent prior distribution.


\subsection{Experiment 2: Map completion in simulation environment} \label{sec:exp2}

\begin{figure}
    \begin{center}
        \frame{\includegraphics[width=8cm]{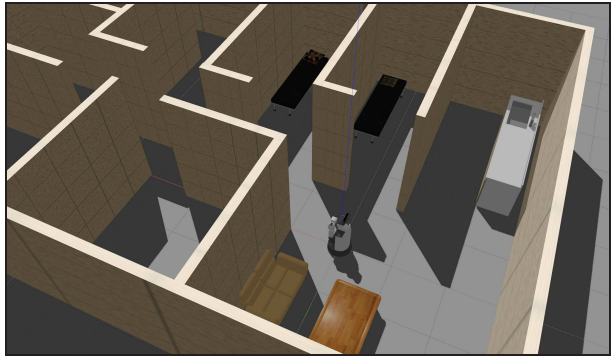}}
        \caption{
            Experiment 2: Example of Gazebo simulation environment used in the experiment.
        }
        \label{fig:gazebo_environment}
    \end{center}
\end{figure}

We evaluate the performance of the proposed method that integrates the network of map completion with SLAM using a simulator environment.
For the experiment, 10 environments were randomly extracted from the test data of the HouseExpo dataset and reproduced in the Gazebo~\cite{gazebo} simulator.
Figure~\ref{fig:gazebo_environment} shows an example of our simulation environment.

The methods to be compared are as follows:
\begin{description}
    \item[(A)] MCN-SLAM (proposed): pix2pix (U-net+discriminator) + Gmapping
    \item[(B)] MCN-SLAM (proposed): U-net + Gmapping
    \item[(C)] Gmapping (grid-based FastSLAM~\cite{Montemerlo2002, Montemerlo2003})
    \item[(D)] Baseline
\end{description}
We compared the results of these methods using only the observations that were acquired without the robot moving.
We use the training data with the highest precision with $m_s$ for the estimated map as a result of the baseline method, same as in Section~\ref{sec:exp1}.

From the result obtained by Gmapping using sufficient observation information as $m_T$, we evaluate the accuracy of the map completion methods by comparing them with $m_T$.
We evaluate the methods in terms of accuracy, precision, recall, and F-measure.

\begin{table}
    \tbl{
        Experiment 2: Experimental results of map completion in Gazebo simulation.
        The top two columns show the experimental results of proposed methods using the HouseExpo dataset for the training data.
        The bottom two columns show the experimental results of proposed methods using the HouseExpo dataset at $T-5$ iteration.
    }{
        \begin{tabular}{l|cccc}
            \toprule
            Method & Accuracy & Precision & Recall & F-measure\\
            \colrule
            (A) MCN-SLAM (proposed): pix2pix & 0.7864 & 0.4486 & \underline{0.7072} & \underline{0.4858}\\
            (B) MCN-SLAM (proposed): U-net & 0.8028 & 0.4686 & \underline{\bf 0.7561} & \underline{\bf 0.5275}\\
            (C) Gmapping & \underline{0.8314} & \underline{0.6836} & 0.0994 & 0.1671\\
            (D) Baseline & \underline{\bf 0.8429} & \underline{\bf 0.7765} & 0.3371 & 0.4062\\
            \colrule
            (A) MCN-SLAM (proposed): pix2pix $(T-5)$ & 0.8715 & \underline{0.6608} & 0.6962 & \underline{0.6546}\\
            (B) MCN-SLAM (proposed): U-net $(T-5)$ & \underline{0.8718} & 0.6306 & \underline{0.7063} & 0.6429\\
            \botrule
        \end{tabular}
    }
    \label{tab:slam:exp1result2}
\end{table}

\begin{figure}
    \begin{center}
        \includegraphics[width=8cm]{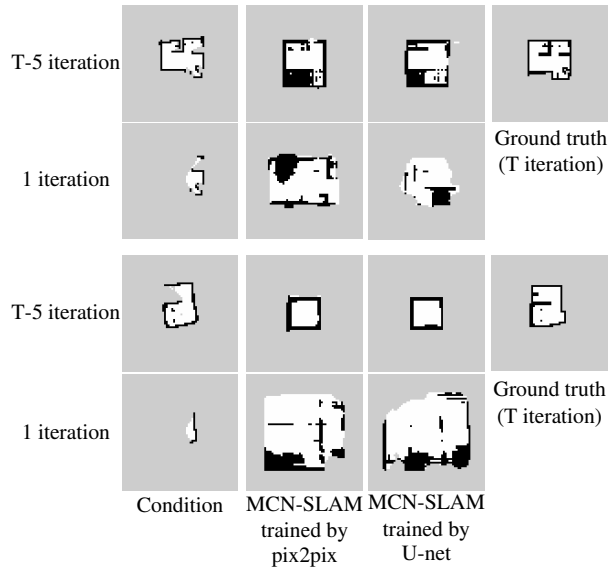}
        \caption{
            Experiment 2: Comparison of the maps generated using pix2pix and U-net as conditions at 1 and $T-5$ iterations of SLAM.
            This comparison shows that the loss by the discriminator is useful to improve the accuracy of map completion when the observation increases and the area of map completion is narrowed.
            The maps of condition show the results of Gmapping. 
        }
        \label{fig:exp1result_network1}
    \end{center}
\end{figure}

Table~\ref{tab:slam:exp1result2} shows the experimental results and Figure~\ref{fig:exp1result_network1} shows an example of generated map images.
When comparing the map estimation methods with and without map completion, the accuracy of the previous methods without map completion was higher, but the F-measure of the proposed method with map completion was higher.
Accuracy has a large effect on unsearched areas, so it is expected that the accuracy is higher in the method that does not use map completion, in which the proportion of unsearched areas is large in the estimation results.
The F-measure of the proposed method using map completion was {1.3 times higher than the baseline}, and it shows that map completion can be performed from limited observations.

The estimation result of the method combining pix2pix with Gmapping was lower in accuracy than using U-net with Gmapping.
From this result, we consider that the discriminator loss is not effective for estimating the environment map using the observation of the initial iteration of SLAM.
In order to investigate in detail the difference depending on the existence of a discriminator, the scores at $T-5$ iteration are shown in the bottom rows of Table~\ref{tab:slam:exp1result2}.
In accuracy, the difference between using U-net and pix2pix is smaller than the result of one iteration, and in F-measure, the result estimated using pix2pix is better.
From this result, it is clear that the discriminator's loss is useful to improve the accuracy of map completion when the observation increases and the area of map completion is narrowed.



\section{Conclusion} \label{conclusion}

We proposed a map completion network-based SLAM (MCN-SLAM) method that incorporates map completion using generative models into the formulation of FastSLAM and does not require observing the entire environment.
Our experiments showed that MCN-SLAM could complement an environment map estimated from few observations and improve the accuracy of the entire map estimation.
{The proposed method was able to transfer structural knowledge to a new environment by acquiring it from many environmental maps via the latent space.}

Improving the accuracy of self-localization by the deep generative model is a challenging task.
In the experiments, we did not verify the effect of the completed map on the robot's self-localization.
In FastSLAM, a map is stored for each particle generated by a particle filter, and a map can be estimated for each particle using a generative model.
By leaving more particles estimating the map with more accurate completion, it is possible to identify a map with incorrect completion better and remove it.
Therefore, an accurately completed map will allow accurate self-localization.
We would like to verify the effect of the above in the future.

In addition, Experiment 1 indicates that the distribution approximated by the map completion network is a domain-dependent prior distribution.
To execute SLAM that performs map completion using the proposed method in a real environment, we can benefit from the observations made by a large number of the same robots in other similar real environments.
This allows performing map estimation while performing map completion in unknown environments using the maps estimated from previously visited environments.

Finally, since our proposed MCN-SLAM method approximates the prior distribution of the environment maps, it is possible to obtain an environment map probabilistically by sampling multiple times.
As a future perspective, MCN-SLAM will be extended as an approach for active exploration of positions where the map prediction is uncertain~\cite{Katyal2019}.
This will hopefully lead to further improvement of robot autonomy.



\section*{Acknowledgments}

This study was partially supported by the Japan Science and Technology Agency (JST) Core Research for Evolutionary Science and Technology (CREST), grant number JPMJCR15E3, and by the Japan Society for the Promotion of Science (JSPS) KAKENHI Grant-in-Aid for Scientific Research (B), grant number 18H03308, and Grant-in-Aid for Scientific Research on Innovative Areas, grant number 16H06569, and Grant-in-Aid for Early-Career Scientists, grant number JP20K19900.

In this paper, we used ``LIFULL HOME'S Dataset'' provided by LIFULL Co., Ltd. via IDR Dataset Service of National Institute of Informatics. 



\bibliographystyle{tADR}
\bibliography{tADR}


\clearpage
\appendix



\section{Experiment 3: Map completion example in a real environment}

\begin{figure}
    \begin{center}
        \includegraphics[width=8cm]{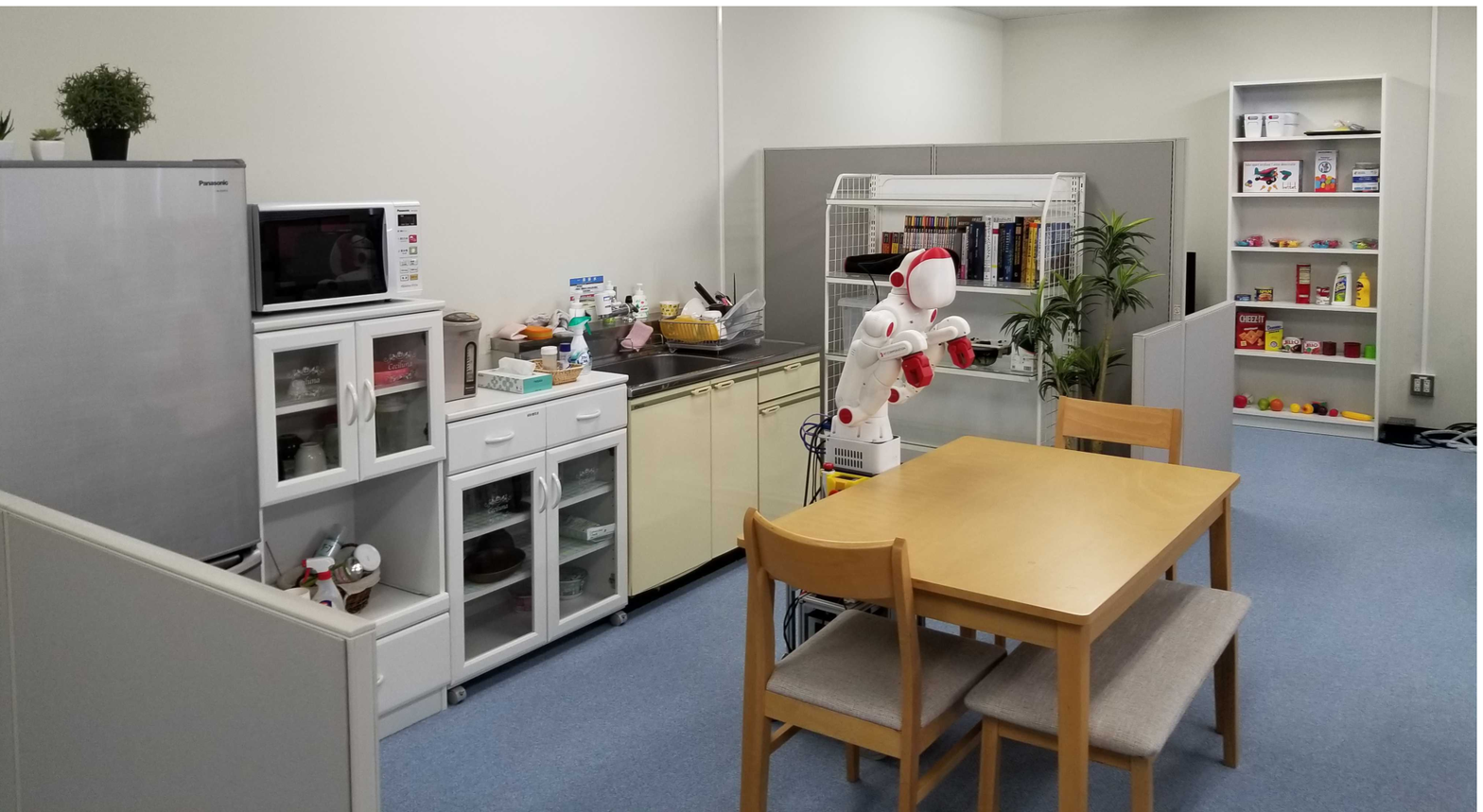}
        \caption{
            Real environment used in the experiment 3.
        }
        \label{fig:real_environment}
    \end{center}
\end{figure}

We perform the map completion using the proposed method and baseline methods in a real environment.
The experimental conditions are the same as in Section~\ref{sec:exp2}.
We use a laboratory room, including an experimental space that imitates a home environment, for the experiment as shown in Figure~\ref{fig:real_environment}.

\begin{figure}
    \begin{center}
        \includegraphics[width=9cm]{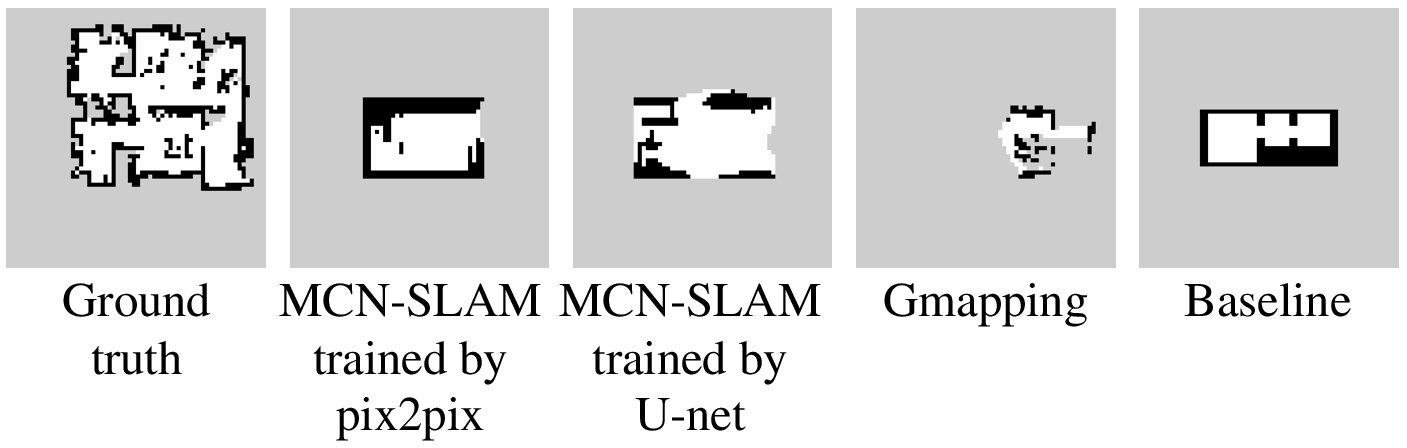}
        \caption{
            Experiment 3: Example of an environment map generated in a real environment by both the proposed and comparison methods.
        }
        \label{fig:exp1result3_1}
    \end{center}
\end{figure}

Table~\ref{tab:slam:exp1result3} and Figure~\ref{fig:exp1result3_1} show the experimental results.
The proposed method obtained better accuracy and F-measure results than the comparison methods from the viewpoint of map completion.
The environment map estimated by the proposed method complements the area where the observation was obtained.
As shown in Figure~\ref{fig:exp1result3_1}, the proposed method succeeded in capturing the rough outline of the walls.
However, when using a network trained with the House Expo dataset prepared by the simulator, it was difficult to capture the complex shapes of the real world because of over-fitting to the domain.
Note that the sensor characteristics of the robot are also different.
As a better way to successfully use the proposed method in real conditions, we are considering training the network using a map of the environment estimated in the real world.
We believe that this problem can be solved in the future by deploying a large number of robots sharing the map estimated in their environment with a cloud-based system.

\begin{table}
    \tbl{
        Experiment 3: Results of the experiment in a real environment.
        The top two columns show the experimental results of proposed methods using HouseExpo as training data.
    }{
        \begin{tabular}{l|cccc}
            \toprule
            Method & Accuracy & Precision & Recall & F-measure\\
            \colrule
            (A) MCN-SLAM (proposed): pix2pix & \underline{\bf 0.7021} & \underline{0.7582} & \underline{0.2507} & \underline{0.3769}\\
            (B) MCN-SLAM (proposed): U-net & \underline{0.7002} & 0.5985 & \underline{\bf 0.3119} & \underline{\bf 0.4101}\\
            (C) Gmapping & 0.6711 & \underline{\bf 0.9817} & 0.1056 & 0.1907\\
            (D) Baseline & 0.6802 & 0.7351 & 0.1945 & 0.3076\\
            \botrule
        \end{tabular}
    }
    \label{tab:slam:exp1result3}
\end{table}

\end{document}